\documentclass[preprint,12pt]{elsarticle}

\usepackage{amsmath}
\usepackage{algorithm}
\usepackage{algorithmic}
\usepackage{multirow}
\usepackage{lscape}
\usepackage{graphicx}
\renewcommand{\vec}[1]{\mathbf{#1}}

\usepackage{amssymb}

\journal{Elsevier}

\begin{document}

\begin{frontmatter}



\title{Currency exchange prediction using machine learning, genetic algorithms and technical analysis}

\author[label1,label2]{Gon\c{c}alo Abreu}
\address[label1]{Instituto de Telecomunica\c{c}\~oes, Instituto Superior T\'ecnico, University of Lisbon, Lisbon, Portugal}
\address[label2]{goncalo.faria.abreu@tecnico.ulisboa.pt}

\author[label1,label3]{Rui Neves}
\address[label3]{rui.neves@tecnico.ulisboa.pt}

\author[label1,label4]{Nuno Horta}
\address[label4]{nuno.horta@tecnico.ulisboa.pt}

\address{}

\begin{abstract}
Technical analysis is used to discover investment opportunities. To test this hypothesis we propose an hybrid system using machine learning techniques together with genetic algorithms. Using technical analysis there are more ways to represent a currency exchange time series than the ones it is possible to test computationally, i.e., it is unfeasible to search the whole input feature space thus a genetic algorithm is an alternative. In this work, an architecture for automatic feature selection is proposed to optimize the cross validated performance estimation of a Naive Bayes model using a genetic algorithm. The proposed architecture improves the return on investment of the unoptimized system from 0,43\% to 10,29\% in the validation set. The features selected and the model decision boundary are visualized using the algorithm t-Distributed Stochastic Neighbor embedding.
\end{abstract}

\begin{keyword}

Machine Learning\sep Genetic Algorithms\sep Naive Bayes\sep Feature Selection\sep Currency Exchange Market\sep t-Distributed Stochastic
Neighbor Embedding




\end{keyword}

\end{frontmatter}


\section{Introduction}
\label{introduction}

The statement that the market follows a random walk is controversial \citep{fama}\citep{lo}. If the market followed a random walk the possibility of profitable traders, professionals that by analyzing the market time series using Technical Analysis (TA) making profit of its predictability and/or inefficiencies, would be impossible. Therefore, if financial market experts with access to this information can consistently beat the market, Machine Learning (ML) methods should be able to capture these patterns (or lack of). This problem can be formulated as a binary classification between an overvalued or undervalued asset (in literature referred to as a mean reversion\citep{poterba} model).

The objective of this work is to investigate the question: is it
possible to build a binary classifier which using TA as input, predicts if the euro-dollar currency exchange is overvalued or undervalued, better than random guessing? And if such a system is possible, a framework to optimize it.

State of the art techniques applied to currency exchange prediction lack focus on system validation, using only a training and validation split, They also  lack embedding techniques  to try to visualize what sort of rules and patterns their architecture has captured, relying on black-box modeling. 

The aspect of system validation will be addressed using multiple ML concepts such as, Cross Validation (CV)\citep{kohavi}. The algorithms used are Naive Bayes and Genetic Algorithms (GA). TA will be used to generate the input features for the model. The main contributions of this work are:

\begin{itemize}
\item Illustrating if TA contains price transformations which can be used to predict the currency exchange time series
\item An architecture for feature selecting using a modified GA
\item Visualization of the model decision boundary and market data structure using the novel t-Distributed Stochastic Neighbor embedding (t-SNE)\citep{maaten}.
\end{itemize}

The remaining part of this paper is organized as follows. Section \ref{Background and Related Work} describes all the fundamental knowledge and related academia works. Section \ref{Proposed Architecture} has an in-depth explanation about the architecture developed for this problem. Section \ref{Architecture Evaluation} presents the results of the architecture, while describing test methodologies and providing a visualization of the embedding technique used, t-SNE. Finally, Section \ref{Conclusion} contains the conclusion.

\section{Background and Related Work}
\label{Background and Related Work}
\subsection{Currency Exchange Market}
\begin{figure}[h]
	\centering
	\includegraphics[width=0.7\linewidth]{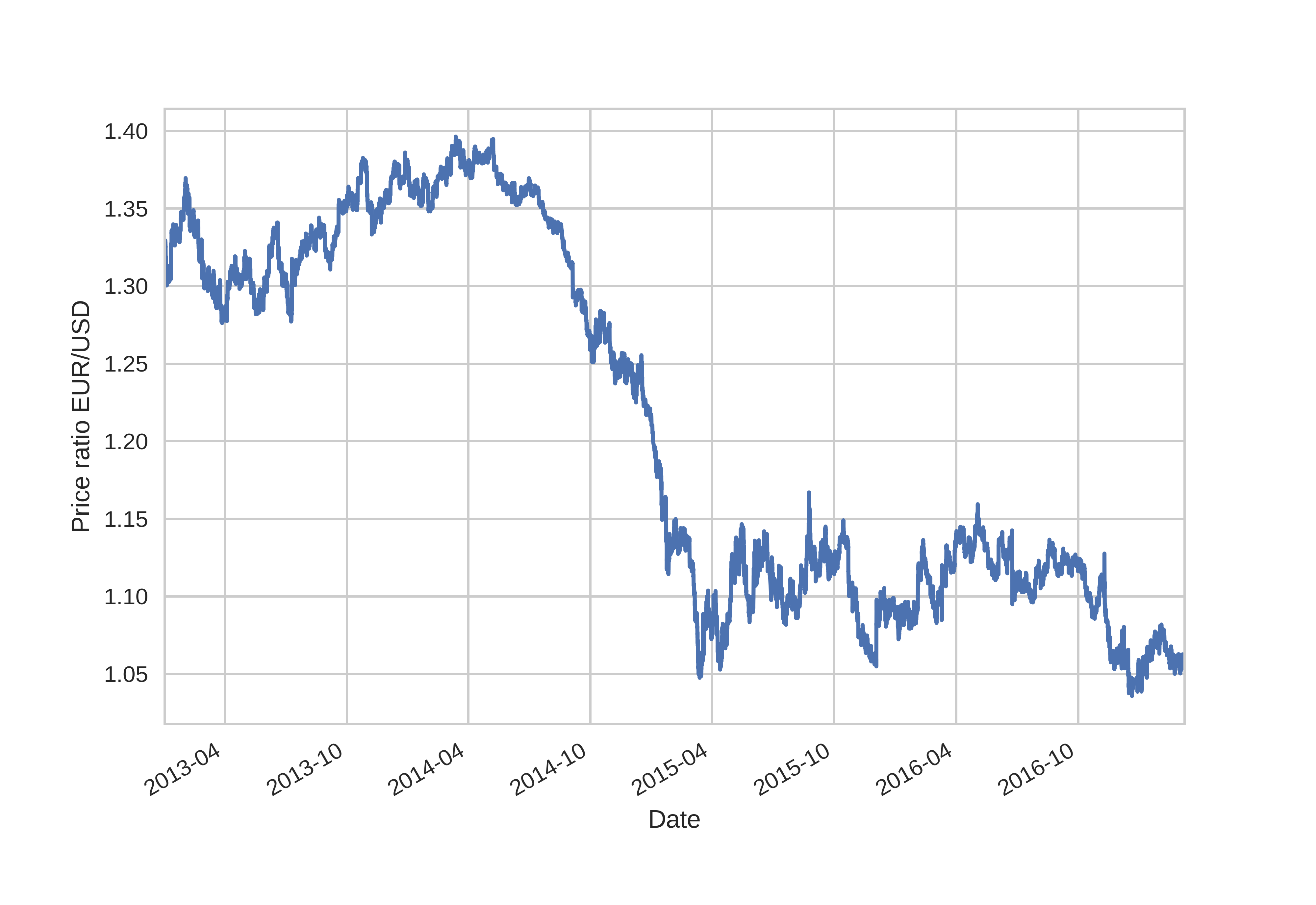}
	\caption{EUR/USD price ratio between January 1st, 2013 and March 9th, 2017}
	\label{fig:EUR_USD}
\end{figure}

The currency exchange market\citep{sarno} is the definition used for the globally decentralized market in which currency pairs are exchanged. In this work, the signal being analyzed is the relationship between the currency trading pair representing the euro-dollar ratio (EUR/USD). A currency trading pair defines the ratio in which euros and dollars will be exchanged at the current market rate, i.e. if the current rate of EUR/USD is 1,18 it means that at this moment if an individual wants to use the market to change currencies from euros to dollars, they will get 1,18 dollars per 1 euro. In Figure \ref{fig:EUR_USD} the EUR/USD time series is presented.

This time series is built from sampling the market at an hourly rate.

\subsection{Technical Analysis}
Although there are two types of analysis used in financial markets, fundamental and technical, only the former one will be used for this work. TA, through the elaboration of mostly price and volume transformations, provides an analysis tool, which financial market experts\citep{pinto} use to forecast a financial instrument in the short-term.

In this work, the following technical indicators were used:
\begin{itemize}
\item Relative Strength Index (RSI) \citep{colby}
\item Commodity Channel Index (CCI) \citep{colby}\citep{kirkpatrick}
\item Moving Average Convergence Divergence (MACD) \citep{achelis}
\item Rate of Change (ROC) \citep{achelis}
\item Stochastic Oscillator \citep{lin}
\item Average True Range (ATR) \citep{kirkpatrick}
\end{itemize}

The literature references for each technical indicator above, contains the formulas used for their calculation.

While analyzing a time series it is necessary to address the problem of non-stationarity\citep{bontempi}. The original currency exchange time-series is a non-stationary process thus, it violates the necessary conditions to apply supervised learning methods, i.e., independent identically distributed samples and having the same training and test data distributions. The set of technical indicators chosen address this problem by turning the original signal into a wide-sense stationary signal (they contain a differencing operation on their formulas)\citep{priestley}\citep{priestley1}.

\begin{figure}[H]
	\centering
	\includegraphics[width=0.7\linewidth]{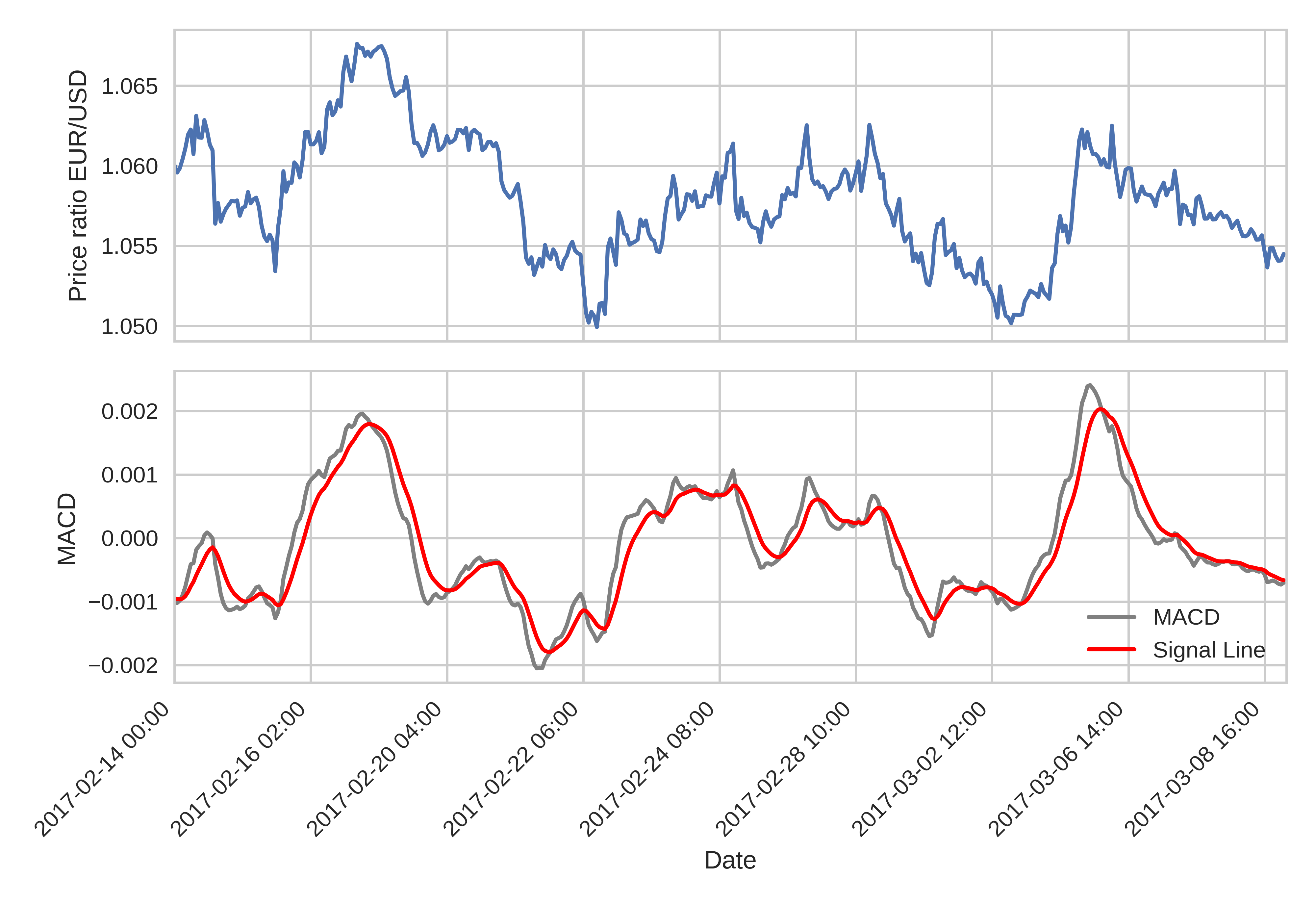}
	\caption{EUR/USD price ratio between a subset of hours with the respective MACD}
	\label{fig:MACD}
\end{figure}

In Figure \ref{fig:MACD} there is an example of the EUR/USD rate with the MACD indicator plotted at the bottom.

\subsubsection{Naive Bayes Classifier}
Naive Bayes classifier is the simplest form of a Bayesian network\cite{zhang} used for classification where all the observed attributes are considered independent. By applying the Bayes theorem to a binary class \(y\) and a dependent vector \( \vec{x}=(x_{1},...,x_{n})\) the resulting relationship is

\begin{equation}
\label{bayes_theorem_eq}
P\left(y|\ \vec{x}\right) = \frac{P\left(\vec{x}|\ y\right) P(y)  }{P(\vec{x})}.
\end{equation}

If a naive assumption that all features are independent is made

\begin{equation}
P\left(x_{i}|\ y,x_{1}, ...,x_{i-1},x_{i+1},...,x_{n}\right) = P(x_{i}|y)
\end{equation}

which for all \(i\) together with Equation (\ref{bayes_theorem_eq})

\begin{equation}
P\left(y|\ \vec{x}\right) = \frac{P(y) \prod_{i=1}^{n} P\left(x_{i}|\ y\right)   }{P(\vec{x})}.
\end{equation}

Considering that \(P(\vec{x})\) is constant given the input, the following classification rule can be built

\begin{equation}
P\left(y|\ \vec{x}\right) \propto P(y) \prod_{i=1}^{n} P\left(x_{i}|\ y\right)  
\end{equation}

which leads to the final relationship, the classification rule

\begin{equation}
y =  \arg\max_{y} P(y) \prod_{i=1}^{n} P\left(x_{i}|\ y\right).
\end{equation}

To use this model a Gaussian assumption is made on the dependent variable using Maximum A Posteriori (MAP) to estimate \(P(y)\) and \(P(x_{i}|y)\)

\begin{equation}
P\left(x_{i}|\ y\right) = \frac{1}{\sqrt{2\pi\sigma_{y}^{2}}}e^{-\frac{(x_{i}-\mu_{y})^2}{2\sigma_{y}^{2}}}
\end{equation}

In this work, a rejection model variant for classification is also be used; for a classification to be accepted, the necessary probability need to be \(>P_{rejection}\),i.e.,

\begin{equation}
y =  \arg\max_{y} P(y) \prod_{i=1}^{n} P\left(x_{i}|\ y\right) \iff \frac{P(y) \prod_{i=1}^{n} P\left(x_{i}|\ y\right)   }{P(\vec{x})} > P_{rejection}
\end{equation}

otherwise the model rejects classifying the given sample.

Naive Bayes has a set of properties which help machine learning optimization tasks, i.e. its complexity is \(\mathcal{O}(NC)\)\cite{murphy}, where \(N\) is the number of features and \(C\) the number of classes, which means its computation footprint is lightweight. This characteristic also makes this algorithm resilient to over-fitting.

\subsubsection{Genetic algorithms}

Genetic algorithms\cite{goldberg} are a family of optimization algorithms originated from the field of evolutionary computation\cite{back} using the Darwinian concepts, such as mutation, reproduction, recombination, and selection.

The most basic element of the algorithm is an individual, which is also called a chromosome. A GA is composed by a populaion of individuals, which are evaluated through the means of a fitness function to have their corresponding fitness attributed. Originally, the chromosome was defined as a bitstring but it can have different representations such as a float/string vector. Each bit/float, in the string/vector, can be referred to, as a gene. The first step of the algorithm is to create an initial population of individuals (chromosomes) and this is done in a random way. This is repeated as often as necessary, to achieve the desired initial population size. After the population is created the fitness of each chromosome is evaluated and assigned. Subsequently, some of the individuals are selected for mating and copied to the mating buffer. In Goldberg's \cite{goldberg} original GA formulation, individuals were selected for mating probabilistically regarding its fitness value, i.e. higher values of fitness translate to higher probability of reproduction. The next concept is entitled crossover (or mutation) and is applied to the population of selected individuals. The mutation operator mutates the gene's inside each chromosome and this is also done in a probabilistic way, for instance, each gene has a probability of being mutated and the mutation follows a rule, usually a continuous probability distribution function. The probability of mutation and the way its performed are implementation decisions. The crossover is the operation which executes the mating of one or more individuals, i.e. it combines bitstrings of different individuals. After the offspring is created it is necessary to fill the rest of the population with individuals, since most GA architectures maintain a fixed size population. There are many strategies for this, one of them is to randomly create new individuals. All these transformations, except the initial creation of the population, are encapsulated in cycles, called generations. The GA should have as many generations as it needs to converge toward a stable solution. Algorithm \ref{alg:ga_pseudo} contains the pseudo code demonstrating what was explained above.

\begin{algorithm}
\caption{GA pseudo code}
\label{alg:ga_pseudo}
\begin{algorithmic}
\STATE generation $n=0$
\STATE initialize population $Pop(n)$ with individuals
\WHILE{fitness criterion not met}
\STATE $n=n+1$
\STATE select a subset of individuals from $Pop(n-1)$ to reproduce, $S(n)$
\STATE cross-over and mutate $S(n)$, resulting on $S'(n)$
\STATE evaluate fitness in $S'(n)$
\STATE create $Pop(n)$ with individuals from $S'(n)$ and $Pop(n-1)$
\ENDWHILE
\end{algorithmic}
\end{algorithm}

\subsubsection{Random immigrants and hyper mutation}

Random immigrants was proposed by Grefenstette\cite{grefenstette}. This alteration replaces the worst elements of the population, at the end of each generation, with randomly generated ones. This procedure is controlled by a hyper-parameter called replacement rate (which is the percentage of the population to be replaced). By doing this, the GA gains the ability of searching different space regions even when it has almost converged, i.e. it gains exhaustive search properties throughout all generations. This can be beneficial to escape converging to a local minima situation. Because the worst elements of the population are the ones being replaced, it doesn't affect the converge rate of the algorithm negatively, considering that the replacement rate is chosen in a conservative way.

Hyper-mutation\cite{cobb} was an alteration to the original GA where the mutation rate changed to a higher level when a trigger is fired. The mutation operation is the responsible GA operator for local search, therefore, when fitness hasn't improved over a set of subsequent generations it considers that it is very close to the maximum fitness, consequently, searching around that solution is the best course of action. Algorithm \ref{alg:ga_immi} contains the pseudo code demonstrating the standard GA with hyper-mutation and random immigrants modification.

\begin{algorithm}
\caption{Random immigrants and hyper-mutation GA pseudo code}
\label{alg:ga_immi}
\begin{algorithmic}
\STATE generation $n=0$
\STATE initialize population $Pop(n)$ with individuals
\WHILE{fitness criterion not met}
\STATE $n=n+1$
\STATE select a subset of individuals from $Pop(n-1)$ to reproduce, $S(n)$
\IF {fitness is not higher than $n-3$ generations ago}
\STATE cross-over and hyper-mutate $S(n)$, resulting on $S'(n)$
\ELSE
\STATE cross-over and mutate $S(n)$, resulting on $S'(n)$
\ENDIF
\STATE evaluate fitness in $S'(n)$
\STATE generate random individuals $R(n)$
\STATE create $Pop(n)$ with individuals from $S'(n)$, $Pop(n-1)$ and $R(n)$
\ENDWHILE
\end{algorithmic}
\end{algorithm}

\subsection{Related Work}
In this subsection, there will be a literature review about the works on currency exchange markets.

\subsubsection{Works on currency exchange markets}
In this section there is an overview of relevant papers which use machine learning techniques to analyze the currency exchange markets, building trading systems.

Das\cite{das} proposed an hybrid algorithm which combined Empirical Mode Decomposition (EDM) and kernel Extreme Learning Machines (KELM) to make daily prediction on a set seven different currency pairs, generating signals for buy, sell and hold. They achieve an annualized return of investment (ROI) of 33,27\%. Their model was made for daily trade and has no reference if leverage was used in the calculations. Kuroda\citep{kuroda} used neural networks(NN) and ensembling strategies together with genetic algorithms to predict the optimal actions for an agent to perform on the USD/JPY currency pair. This system achieved 15,79\% of annualized return. Petropoulos\citep{petropoulos} used genetic algorithms together with multiple machine learning algorithms to analyze correlations across multiple currency pairs to generate signals for daily trading. The best annualized return achieved using leverage is 17,4\%. Jubert de Almeida\citep{dealmeida} proposed a system based on genetic algorithms optimizing a support vector machine model for the EUR/USD currency pair. Without the leverage approach the system achieved a 8,9\% annualized return. Sidehabi\citep{sidehabi} proposed a system based on statical methods together with machine learning to forecast the EUR/USD currency pair. In this work, no trading system is proposed although the performance of the model is measured with the cost metric root mean squared error (RMSE) achieving a mean value of 0,001322. Deng\citep{deng} proposed a system combining multiple kernels learning for regression (MKR) together with a genetic algorithm to construct a set of trading rules. The mean annualized return using leverage was 15,4\%. Evans\citep{evans} combined NN with genetic algorithms to build an intra-day trading algorithm for currency exchange markets. The performance was evaluated on three base currencies using leverage, achieving a mean annualized return of 27,83\%. Georgios\citep{georgios} used NN and radial basis functions (RBF) together with particle swarm optimization with a time varying leverage trading strategy. The achieved result was a mean annualized return of 25,01\%. Table \ref{bibtable} contains a summary of the works reviewed.

\begin{table}[]
\centering
\caption{Works on currency exchange markets summary}
\label{bibtable}
\begin{center}
 \resizebox{\linewidth}{!}{%
 \begin{tabular}{|c c c c c c|} 
 \hline
 Reference & Date & Approach & Evaluation Metric & Leverage & Metric Value \\ [0.5ex] 
 \hline\hline
 \citep{das} & 2017 & EDM, KELM & Annualized ROI & - & 33,27\% \\ 
 \hline
 \citep{kuroda} & 2017 & NN & Annualized ROI & - & 15,79\% \\
 \hline
 \citep{petropoulos} & 2017 & GA and multiple classifiers & Annualized ROI & Yes & 17,4\% \\
 \hline
 \citep{dealmeida} & 2018 & GA, SVM & Annualized ROI & No & 8,9\% \\
 \hline
 \citep{sidehabi} & 2016 & ML and statistical methods & RMSE & - & 0,001322 \\  
 \hline
 \citep{deng} & 2015 & MKR, GA & Annualized ROI & Yes & 15,4\% \\  
 \hline
 \citep{evans} & 2013 & NN, GA & Annualized ROI & Yes & 27,83\% \\  
 \hline
 \citep{georgios} & 2013 & NN, GA and RBF & Annualized ROI & Yes & 25,01\% \\ [0.5ex]
 \hline
\end{tabular}}
\end{center}
\end{table}

\section{Proposed Architecture}
\label{Proposed Architecture}

In this work a classifier is built to predict whether the EUR/USD time series is overbought or oversold, generating buy and sell signals which will serve as input for a trading system. The inputs of the classifier are technical indicators.

\subsection{Overview}

From the EUR/USD time series, a set of technical
indicators are calculated. The baseline Naive Bayes model will have as input features, the technical indicators calculated with the parameters that the literature suggests has default values. This model will be compared to one in which the features and their parameters are selected through GA optimization. After the training process, both methodologies performance is estimated by cross-validation, evaluated through the use of a held out validation set, and compared.

\subsection{Binary Target Formulation}

When using methods of supervised learning it is necessary to formulate the quantity which is going to be predicted.  The binary target is going to be represented as \(y_{t}\). \(\vec{y} = (y_{t_{1}},...,y_{t_{n}})\)  follows a binomial probability distribution, i.e. \(y \in \{0,1\}\) , where 1 symbolizes if the signal had a positive variation and 0 a negative one. If \(Close_{t}\) is the closing price of the EUR/USD rate at the current hour and \(Close_{t-1}\) the closing price in the previous hour, then \(y_{t}\) can be defined as,

\begin{equation}
y_{t} = 
     \begin{cases}
       1&: Close_{t}-Close_{t-1} \geq 0\\
       0&: Close_{t}-Close_{t-1} < 0\\
     \end{cases}
\end{equation}

The trading rule used follows this target formulation, i.e. if the model predicts a variation so that $P\left(y|\ \vec{x}\right) > P_{rejection}$ an order on the market is simulated, and closed after an hour (the sampling rate of the time-series).

\subsection{Optimized features through GA search}
Each technical indicator has one or many free parameters which affect the way they filter the time series, hence it is impossible to know a priori which parameters translate to the best performance accuracy of the classifier. Likewise, since almost every technical indicator involves the calculation of a moving average with regards to a certain time window, some of them might overlap and be redundant (for instance RSI and CCI have the same default window which they use to calculate the moving average of the given time series). There is a combination of features (and feature parameters) which will result in the best overall accuracy of the system, but since there are more possible combinations (\(>10^{21}\) possibilities) than the computational power available to try a purely exhaustive search, a modified GA, with random immigrants and hyper mutation approach is proposed to search the feature space efficiently.

\begin{figure}[h]
	\centering
	\includegraphics[width=0.75\linewidth]{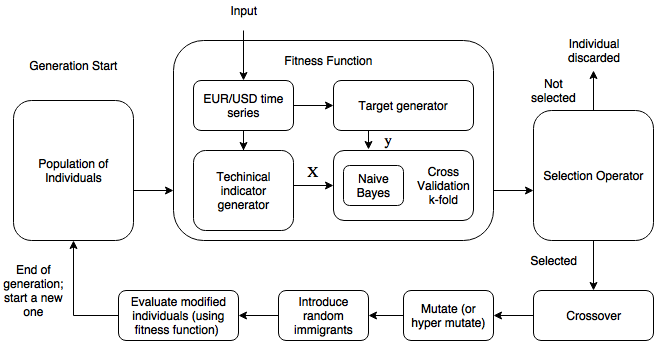}
	\caption{Proposed architecture pipeline}
	\label{fig:ga_pipeline}
\end{figure}

In Figure \ref{fig:ga_pipeline} a diagram of the proposed architecture pipeline is shown. 

Given a binary classifier and a time series dataset, the GA is going to search and optimize the set of technical indicators which maximizes the accuracy of the given classifier. The output of the system, not only gives the best subset of technical indicators (and its parameters), but also makes an estimation, with confidence intervals, on what the performance of the system on unseen data will be. CV is used to prevent over-fitting and estimate the accuracy confidence intervals. For further confirmation, the system will be validated using a validation set, unseen during the optimization process.

\subsection{Training and validation dataset splits}

To have reliable performance estimation a split in the time series is necessary, dividing it in a training and validation set. The training set will be used to generate models and estimate their future performance. The validation will always be held out, until the training process is finished and the performance confidence intervals estimated. After this step, the system performance is validated on a held-out validation set. Although CV estimation is considered reliable, as a performance estimation, having a held-out validation set is an extra layer of system validation. The EUR/USD dataset contains all the time series information between the period from January 1st, 2013 till March 9th, 2017 sampled at an hourly rate. The training set will be the first 80\% of the time series, leaving the validation set with the final 20\%(according to the Pareto principle\citep{suthaharan}).

\begin{figure}[H]
	\centering
	\includegraphics[width=0.75\linewidth]{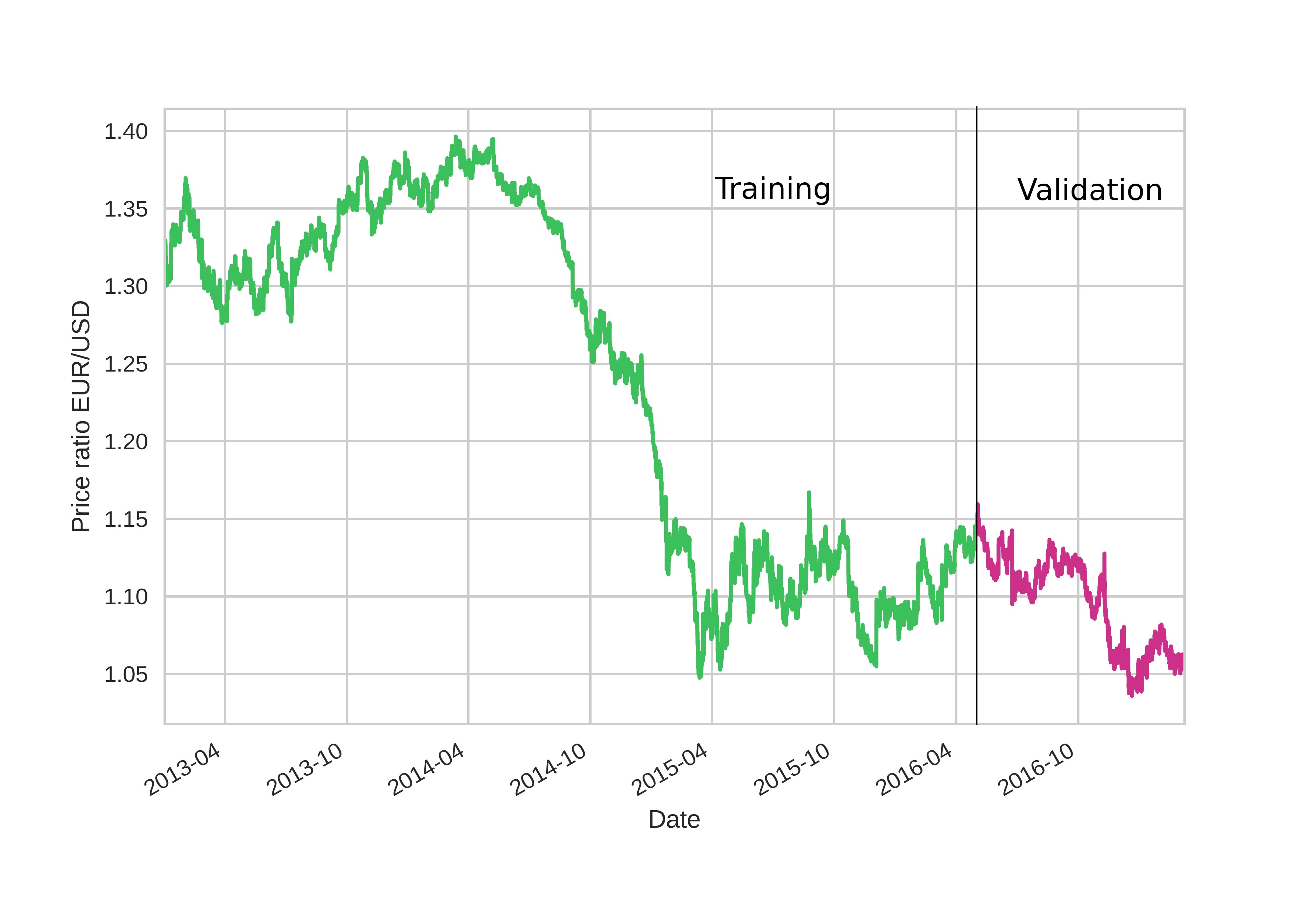}
	\caption{EUR/USD price rate divided in training and validation splits}
	\label{fig:EUR_USD_train_test}
\end{figure}

In Figure \ref{fig:EUR_USD_train_test} there is a visual representation of this split. By making this split, the train set contains 20726 training samples leaving the validation set with 5182 samples.

\section{Architecture Evaluation}
\label{Architecture Evaluation}

The proposed architecture is evaluated against a baseline model with no optimization. Section \ref{baseline_model} refers to the baseline model so a latter comparison can be made. The ML scoring metrics used are accuracy, precision and recall\citep{davis}.

\subsection{Naive Bayes classifier with rejection}
\label{baseline_model}

The first step is to estimate the performance using \textit{k}-fold CV scheme. The estimated accuracy for a 7-fold CV scheme is 53,4\% \( \pm 2,2\% \) (a mean estimator would yield 50,1\% accuracy). To enhance the accuracy of the simple binary model, rejection was implemented so that some samples are rejected if the model is too uncertain about what class they belong to, i.e. \(P_{rejection}\) was chosen so that half the samples are rejected. Although there are areas where rejecting samples is not optimal, in a financial trading model it is beneficial to implement rejection, to mitigate the high risk of misclassifying. The rejection was chosen, so that half of the samples are rejected. In Table \ref{naive_rejection} the Naive Bayes classifier metrics on training and validation sets are presented.

\begin{table}[ht]
\centering
\caption{Training and validation results for the Naive Bayes classifier with rejection}
\label{naive_rejection}
\resizebox{\linewidth}{!}{%
\begin{tabular}{|l|l|l|l|l|l|l|}
\hline
\multicolumn{1}{|c|}{\multirow{2}{*}{\textbf{Label}}} & \multicolumn{3}{c|}{\textbf{Training}} & \multicolumn{3}{c|}{\textbf{Validation}} \\ \cline{2-7} 
\multicolumn{1}{|c|}{}                                & Precision    & Recall     & ROI        & Precision   & Recall    & ROI      \\ \hline
0                                                     & 54,58\%      & 28,71\%    & 18,52\%    & 52,75\%     & 25,48\%   & -2,17\%  \\ \hline
1                                                     & 54,92\%      & 25,35\%    & 10,01\%    & 50,03\%     & 27,90\%   & 2,60\%   \\ \hline
Average/Total                                         & 54,75\%      & 27,93\%    & 28,53\%    & 51,39\%     & 26,69\%   & 0,43\%   \\ \hline
\end{tabular}}
\end{table}

\begin{figure}[h]
	\centering
	\includegraphics[width=0.75\linewidth]{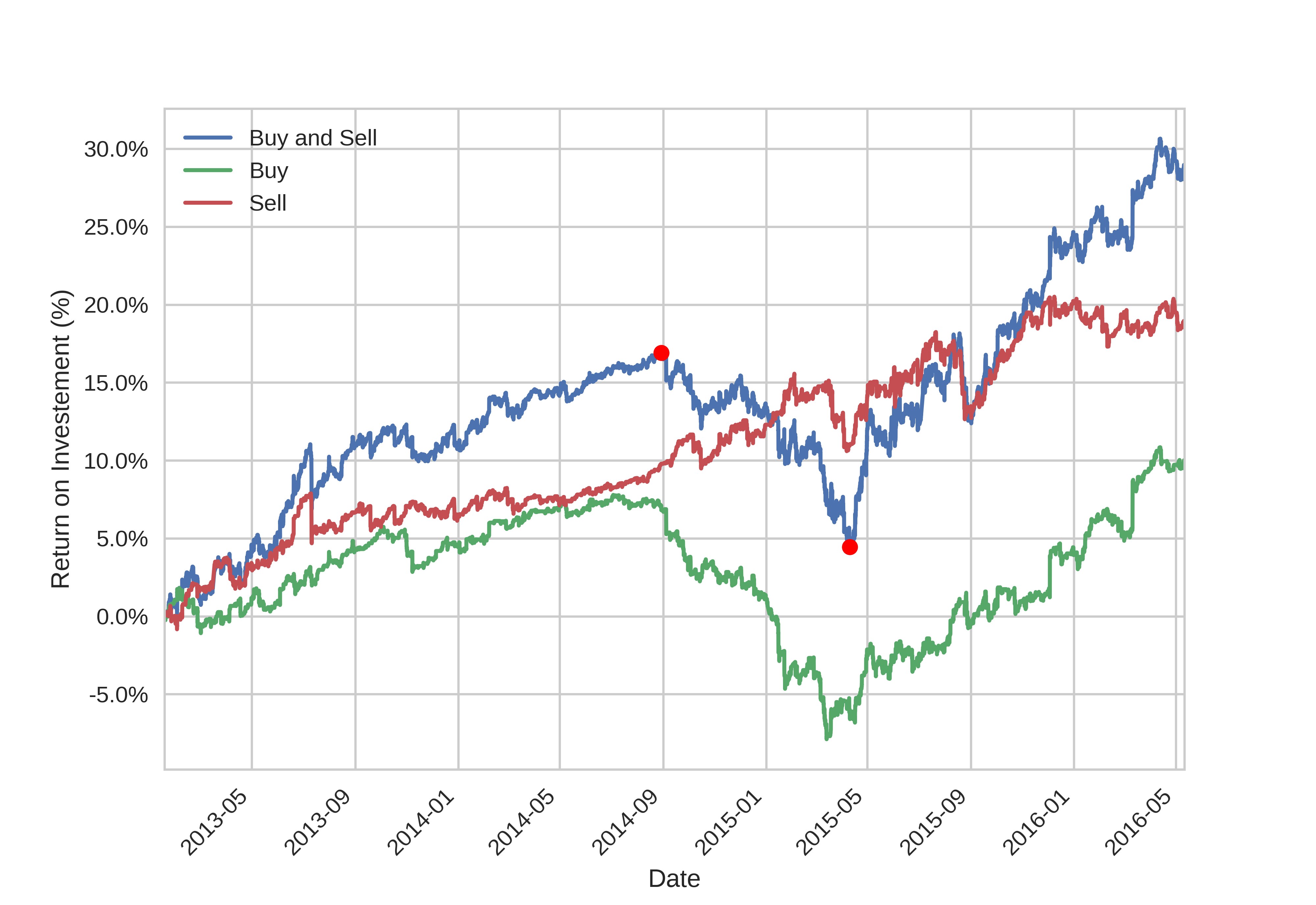}
	\caption{Market simulation on the rejection model with no optimization using the training set with predictions generated through CV}
	\label{fig:naive_bayes_train_reje}
\end{figure}

The market simulation is performed with the rejection model giving signals for buy and sell positions, with no leverage. In Figure \ref{fig:naive_bayes_train_reje} the performance of market simulation in the training set can be seen, the predictions used to make this plot were generated through \textit{k}-fold CV to prevent training over-fit. The model would have a positive result yielding approximately 28,53\% returns over the course of 3 years (9,51\% annualized return).

The "Buy and Sell" line accounts for the summed result of both buy and sell orders. The max drawdown period is between the two red dots (September 2014 and April 2015), and its value is -12,44\%.

\begin{figure}[h]
	\centering
	\includegraphics[width=0.75\linewidth]{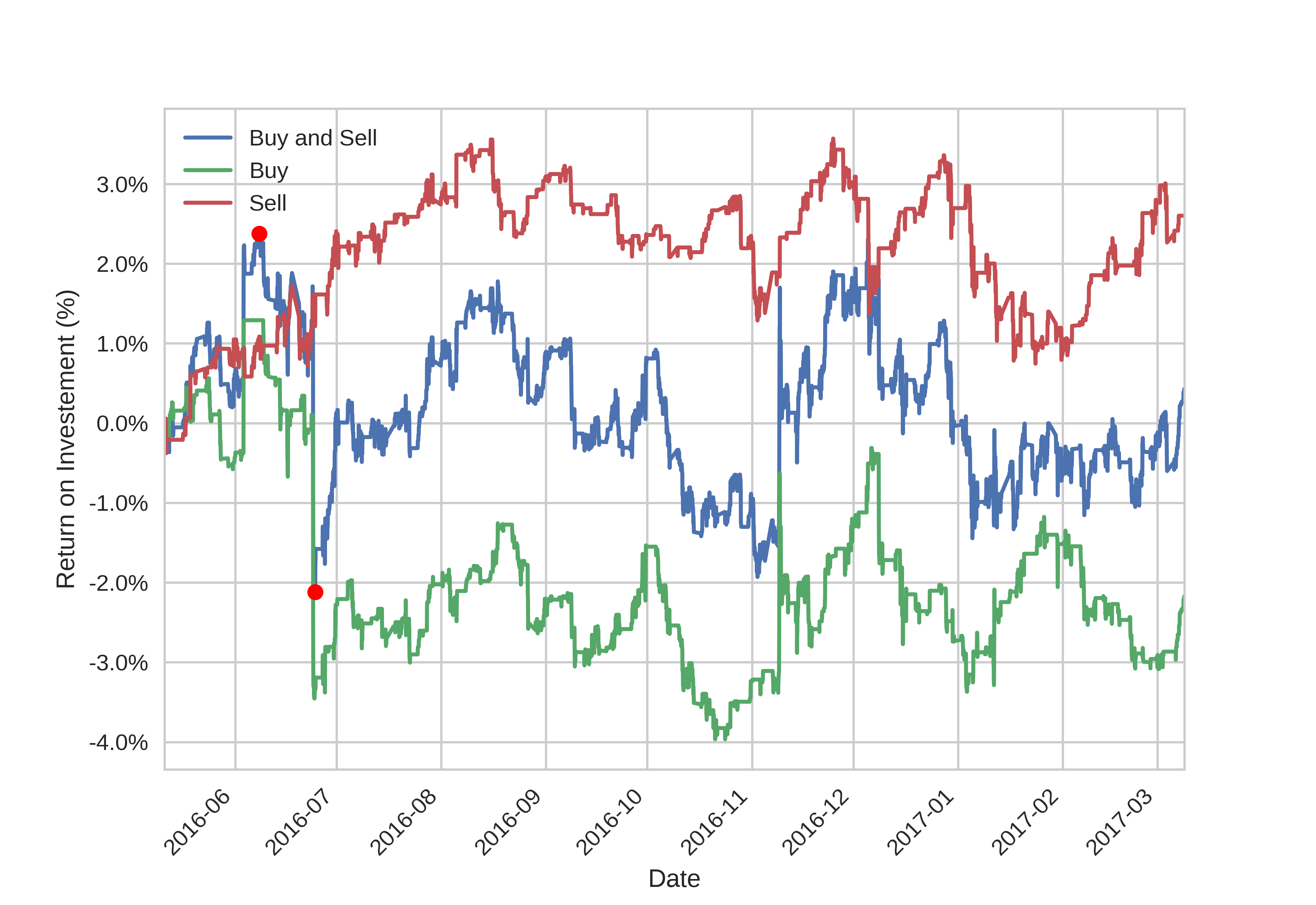}
	\caption{Market simulation on the rejection model with no optimization using the validation set}
	\label{fig:naive_bayes_test_reje}
\end{figure}

The performance of the model in the validation set is presented on Figure \ref{fig:naive_bayes_test_reje}. Since the validation set was never seen in the training process, it is a good indicator that the problem is well formulated and result estimations work as expected. This model would yield approximately 0,43\% (0,47\% annualized return) through the course of 11 months. The max drawdown is between the two red dots in the plot (June 2016 and February 2017), and its value is -4,49\%.

\subsection{Naive Bayes classifier with rejection and features optimized by the proposed GA}

The architecture developed for this work will optimize the simple Naive Bayes model input features, by searching its the optimal combination and parameters. Since the GA performs searches based on random sampled individuals to validate the configuration parameters, the convergence of multiple runs must be inside an acceptable interval

\begin{figure}[h]
	\centering
	\includegraphics[width=0.75\linewidth]{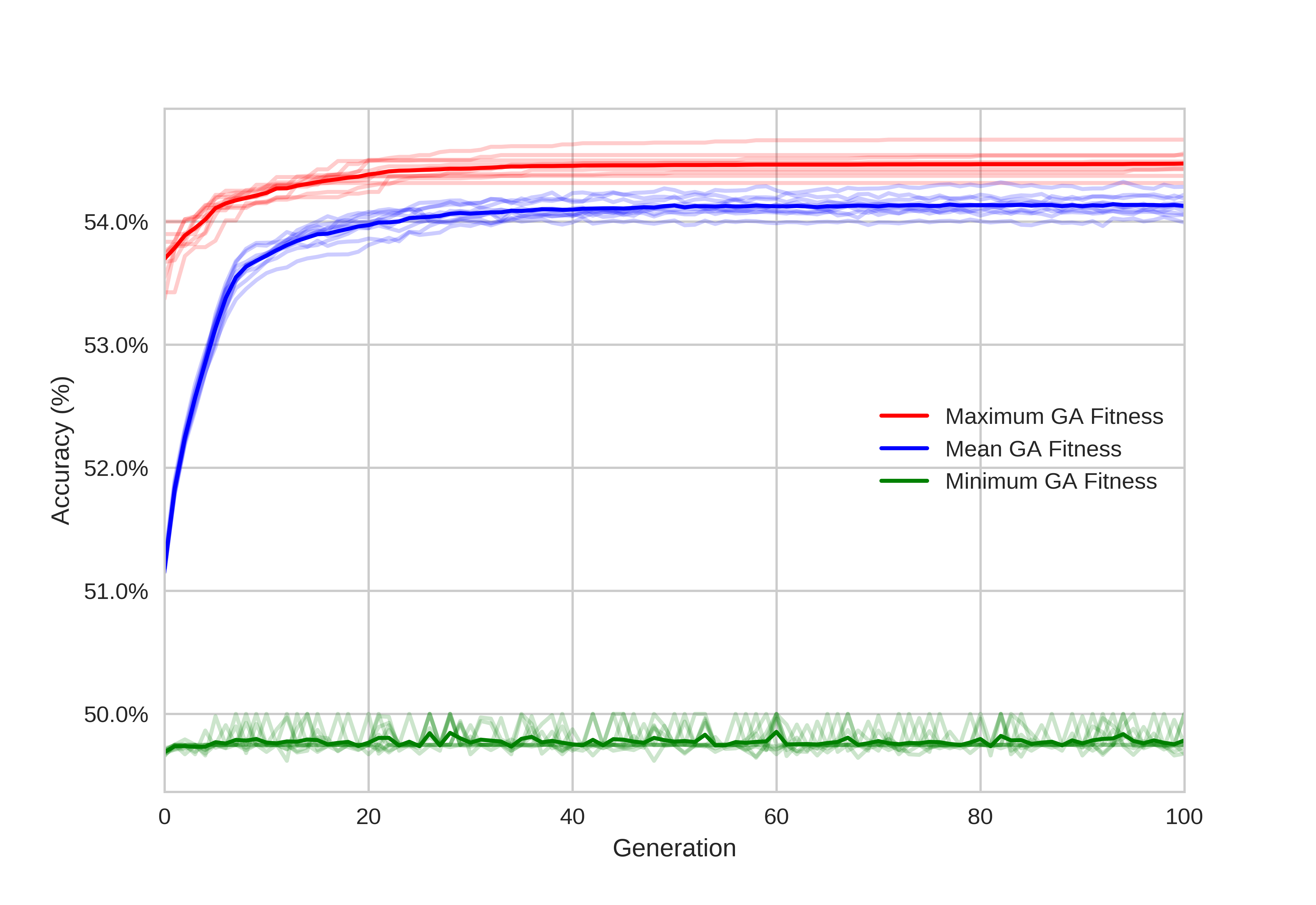}
	\caption{Fitness value across 10 different GA runs (the accuracy is estimated though 7-fold CV)}
	\label{fig:convergence_runs}
\end{figure}

The convergence analysis of the GA was made using 10 different runs of the algorithm. On Figure \ref{fig:convergence_runs} the maximum, average, and minimum GA fitness across all the runs is shown (solid lines represent the mean of all runs, while the shaded, each individual run).

To make a fair comparison with the previous case study, \(P_{rejection}\) was chosen so that the proposed optimized classifier rejected half the samples. The best CV estimated accuracy achieved by a GA was 54,7 \% \( \pm 2,2\% \) . The detailed metrics of the optimized binary classifier in training set are present in Table \ref{naive_rejection_opti}.

\begin{table}[h]
\centering
\caption{Training and validation results for the Naive Bayes classifier with rejection and feature optimized by the GA}
\label{naive_rejection_opti}
\resizebox{\linewidth}{!}{%
\begin{tabular}{|l|l|l|l|l|l|l|}
\hline
\multicolumn{1}{|c|}{\multirow{2}{*}{\textbf{Label}}} & \multicolumn{3}{c|}{\textbf{Training}} & \multicolumn{3}{c|}{\textbf{Validation}} \\ \cline{2-7} 
\multicolumn{1}{|c|}{}                                & Precision    & Recall     & ROI        & Precision   & Recall    & ROI      \\ \hline
0                                                     & 55,77\%      & 26,15\%    & 21,63\%    & 55,81\%     & 24,79\%   & 7,12\%   \\ \hline
1                                                     & 56,15\%      & 29,05\%    & 23,13\%    & 52,09\%     & 29,49\%   & 3,17\%   \\ \hline
Average/Total                                         & 55,96\%      & 27,60\%    & 44,76\%    & 53,95\%     & 27,14\%   & 10,29\%  \\ \hline
\end{tabular}}
\end{table}

It is possible to conclude, through the analysis of Table \ref{naive_rejection} and Table \ref{naive_rejection_opti} that the proposed optimization achieves better results.

\begin{figure}[h]
	\centering
	\includegraphics[width=0.75\linewidth]{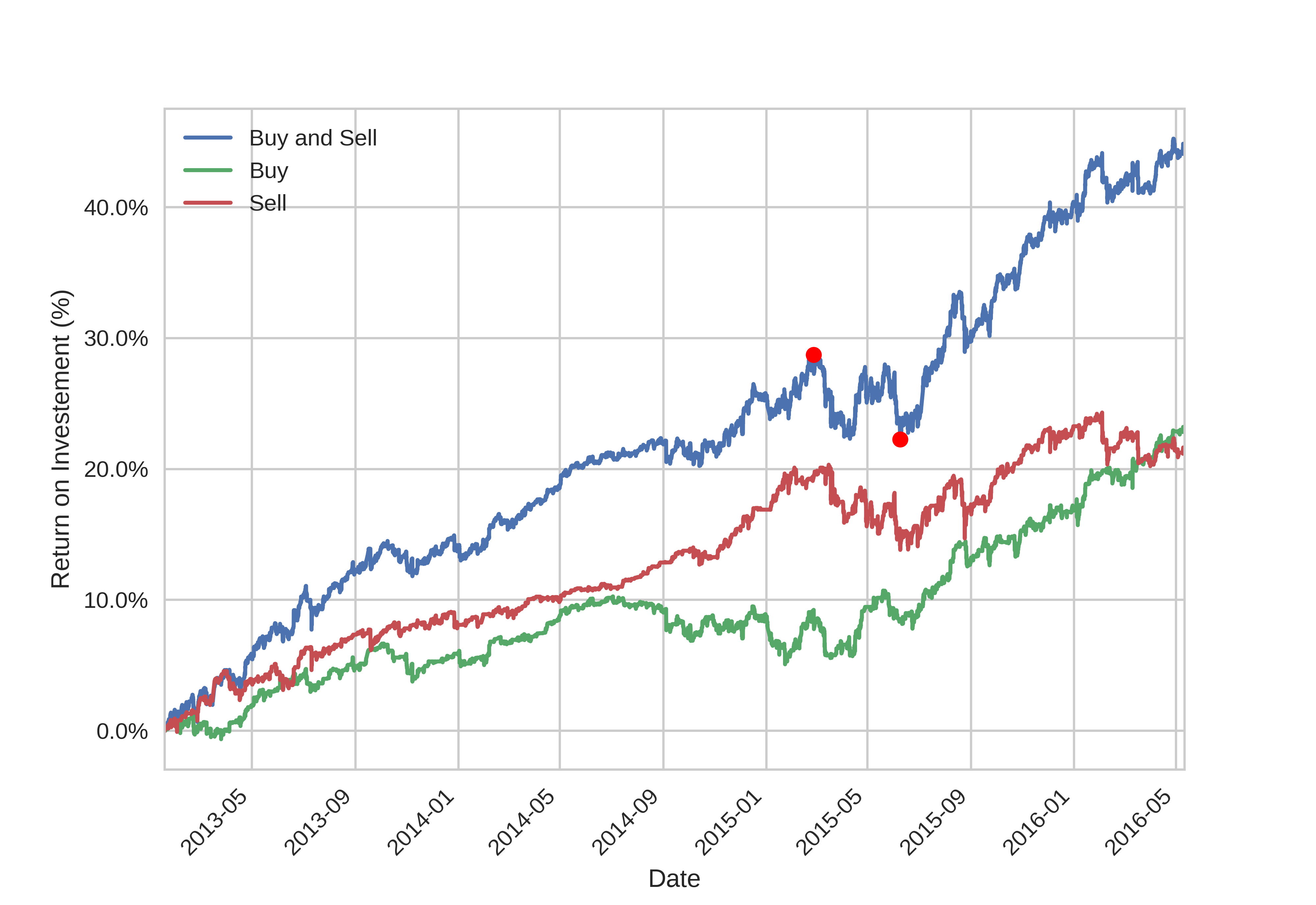}
	\caption{Market simulation on the rejection model with feature optimization using the training set with predictions generated through CV}
	\label{fig:opti_reje_train}
\end{figure}

The best individual optimized through the GA with Naive Bayes rejection algorithm yielded 44,76\% (14,92\% annualized) in returns on the market simulation (the unoptimized model had 28,96\% under the same conditions). The max drawdown period is shown between the two red dots, and its value is -6,46\%. The detailed results are illustrated in Figure \ref{fig:opti_reje_train}.

\begin{figure}[H]
	\centering
	\includegraphics[width=0.75\linewidth]{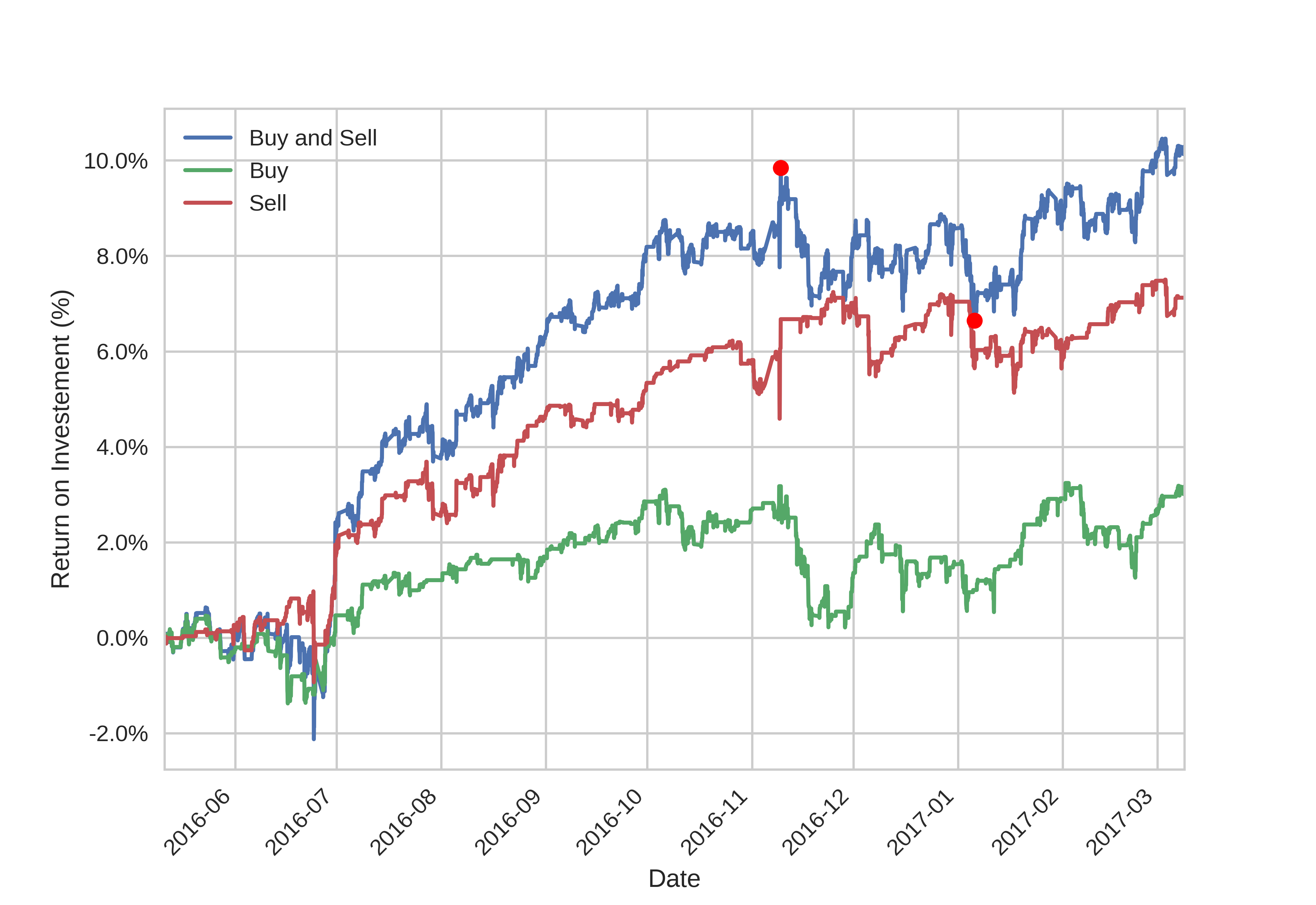}
	\caption{Market simulation performance on the held-out validation set of the rejection model with feature optimization using GA}
	\label{fig:opti_reje_test}
\end{figure}

In Figure \ref{fig:opti_reje_test}, the performance of the market simulation of the optimized model in the validation set is presented. The return on investment is 10,29\% (11,23\% annualized), which is a significant improvement when comparing to the return on investment achieved by the unoptimized model, 0,43\%. The max drawdown, represented between the two red dots is -3,20\%.

\subsection{Model and data visualization with t-SNE}

One relevant question to answer while developing an ML application is: "what sort of relationships and patterns are encoded within the model and data". Recently there have been some advances on data visualization with the development of embedding algorithms such as t-SNE\citep{maaten}.

t-SNE is a visualization technique for laying out a large and high dimensional datasets into 2-d maps while preserving the local structure in the data. A brief explanation of the algorithm is the following: it minimizes an objective function, using gradient descent, which measures the discrepancy between similarities in high dimensional data, projecting those onto a lower dimensional map (hence the name embedding). To summarize, t-SNE is used to build a 2-d map which represents local similarities of a higher dimensional map while minimizing the Kullback-Leibler divergence between the two. t-SNE favors local similarity between the high dimensional points, preserving it, i.e. distance between non-similar points on the 2-d map may have a non-linear interpretation.

One recent paper by authors working at Goggle DeepMind\citep{mnih}, "Human-level control through deep reinforcement learning" uses t-SNE to show state values of what the maximum expected reward will be, according to the action performed.

\begin{landscape}
\begin{figure}[h]
	\centering
	\includegraphics[width=\linewidth]{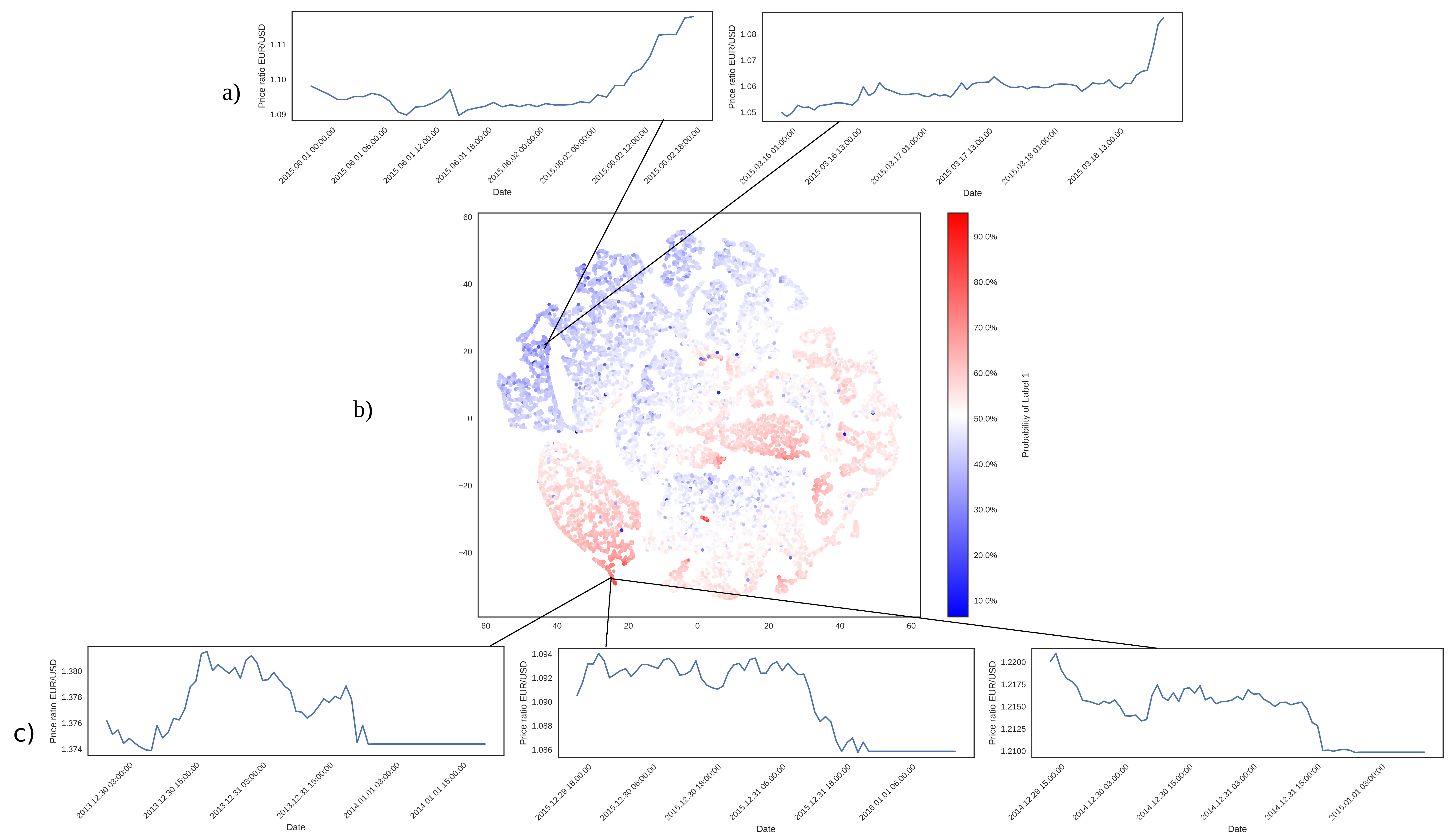}
	\caption{a) Market observations with a high probability of having a negative variation in the next time period. b) t-SNE embedding, generated with the input features discovered by the proposed architecture, where each sample (point in space) represents a market observation. The color gradient is the probability, of the next time period having a positive market variation. c) Market observations with a high probability of having a positive variation in the next time period (the beginning of year 2014, 2015 and 2016).}
	\label{fig:tsne}
\end{figure}
\end{landscape}

Figure \ref{fig:tsne} is an attempt to visualize what sort of patterns, the model developed for this work has learned. The input features used were the ones discovered and optimized by the GA. The color gradient is the probability, of the next time period having a positive market variation, calculated by the Naive Bayes binary classifier using a \textit{k}-fold CV scheme. Each point in Figure \ref{fig:tsne} is a market time series observation, where in each cluster, observations that are near each other (preserving local structure) are expected to be similar (remember that each market observation is represented by multiple technical indicators, thus a high dimensional feature map can be embedded by the t-SNE onto a 2-d map). In some regions of this plot, similar points have equal probability of being label 1 (or label 0), represented by the colors red and blue respectively, e.g on the bottom three price charts, each plot represents the beginning of year 2014, 2015 and 2016; Naive Bayes model learned this pattern and assigns it with a higher probability of having a positive variation (even though the model has no time awareness, just pattern awareness through the use of technical indicators). There are some market observations which are similar and have more probability of being of a certain class, and the most certain ones are market inefficiencies. For instance, both price charts on the top have a lower probability of having a positive market variation, thus having a high probability of a negative market variation, furthermore it is interesting to note that they both show a similar behavior, an abnormal increase on the EUR/USD price rate in a short period of time. It is an anomalous market behavior to have such price variation in a short amount of time, and the Naive Bayes model predicts that there is a higher probability of a negative variation (mean reversion). If the market followed a true random walk distribution there shouldn't be plot areas where the probability of having a positive variation, or not, is much higher, or much lower, than 50\%, since the mean of observed variation is $\approx50\%$. It appears that the developed model is capturing some memory (and/or repeated pattern) present in the signal variation represented in the time series.

Since Figure \ref{fig:tsne} is built using an unsupervised technique, without knowing what the target is, it would be interesting to do a thorough study on what the local clusters might mean, thus building new and more representative features. Plotting the color gradient shows there is an important correlation between what the model is capturing and the local structure present in the dataset.

\section{Conclusion}
\label{Conclusion}
Evolutionary computation mixed with probabilistic classifiers  provide a simple and efficient infrastructure to search and optimize features. The currency exchange market, specifically the EUR/USD, was the dataset in which the proposed architecture was tested. Although the initial objective for this work was to prove that there is a benefit of mixing different areas of ML, to achieve results in an efficient manner, the experiments developed to test this, provide interesting conclusions about the currency exchange market. Although the Naive Bayes rejection classifier already achieved better performance than random guessing, the proposed architecture, which searched and optimized features using a modified GA, proved that it is possible to boost the Naive Bayes classifier accuracy by a considerable amount (from 51,39\%, on validation set, to 53,95\%, which translates to a significant improvement on the trading system, 0,43\% to 10,29\% ROI). Model visualization is topic with growing interest on the ML community to detach from black-box models with no interpretability. By making use of t-SNE algorithm it is possible to visualize what type of patterns the Naive Bayes is learning from the input data. This visualization opens the path to future work since it has revealed a lot of local clusters which were not present in the standard technical indicator default parameters, only in the GA optimized ones. It is possible to conclude that the points where the algorithm is most certain of the price direction are based on mean reversions after a market inefficiency, such as an abnormal price variation in a short amount of time.

\section*{Acknowledgments}
This work was supported in part by Funda\c{c}\~ao para a Ci\^encia e a Tecnologia (Project UID/EEA/50008/2013).




\clearpage
\section{References}
\begin{thebibliography}{00}


\bibitem[Fama(1995)]{fama}
Fama, Eugene F. "Random walks in stock market prices." Financial analysts journal 51.1 (1995): 75-80.

\bibitem[Lo(1988)]{lo}
Lo, Andrew W., and A. Craig MacKinlay. "Stock market prices do not follow random walks: Evidence from a simple specification test." The review of financial studies 1.1 (1988): 41-66.

\bibitem[Poterba(1988)]{poterba}
Poterba, James M., and Lawrence H. Summers. "Mean reversion in stock prices: Evidence and implications." Journal of financial economics 22.1 (1988): 27-59.

\bibitem[Kohavi(1995)]{kohavi}
Kohavi, Ron. "A study of cross-validation and bootstrap for accuracy estimation and model selection." Ijcai. Vol. 14. No. 2. 1995.

\bibitem[Maaten(2008)]{maaten}
Maaten, Laurens van der, and Geoffrey Hinton. "Visualizing data using t-SNE." Journal of Machine Learning Research 9.Nov (2008): 2579-2605.

\bibitem[Sarno(2001)]{sarno}
Sarno, Lucio, and Mark P. Taylor. "Official intervention in the foreign exchange market: is it effective and, if so, how does it work?." journal of Economic Literature 39.3 (2001): 839-868.

\bibitem[Pinto(2015)]{pinto}
Pinto, J. M., Neves, R. F., \& Horta, N. (2015). Boosting Trading
Strategies performance using VIX indicator together with a dual-objective
Evolutionary Computation optimizer. Expert Systems with Applications,
42(19), 6699-6716.

\bibitem[Colby(1988)]{colby}
Colby, Robert W., and Thomas A. Meyers. The encyclopedia of technical
market indicators. New York: Irwin, 1988.

\bibitem[Kirkpatrick(2010)]{kirkpatrick}
Kirkpatrick II, Charles D., and Julie A. Dahlquist. Technical analysis: the
complete resource for financial market technicians. FT press, 2010.

\bibitem[Achelis(2001)]{achelis}
Achelis, Steven B. Technical Analysis from A to Z. New York: McGraw
Hill, 2001.

\bibitem[Bontempi(2013)]{bontempi}
Bontempi, Gianluca. "Machine learning strategies for time series prediction." European Business Intelligence Summer School, Hammamet, Lecture (2013).

\bibitem[Priestley(1981)]{priestley}
Priestley, Maurice Bertram. "Spectral analysis and time series." (1981).

\bibitem[Priestley(1988)]{priestley1}
Priestley, Maurice Bertram. "Non-linear and non-stationary time series analysis." (1988).

\bibitem[Lin(2011)]{lin}
Lin, Xiaowei, Zehong Yang, and Yixu Song. "Intelligent stock trading
system based on improved technical analysis and Echo State Network."
Expert systems with Applications 38.9 (2011): 11347-11354.

\bibitem[Zhang(2004)]{zhang}
Zhang, Harry. "The optimality of naive Bayes." AA 1.2 (2004): 3.

\bibitem[Murphy(2012)]{murphy}
Murphy, Kevin P. Machine learning: a probabilistic perspective. MIT press, 2012.

\bibitem[Booker(1989)]{goldberg}
Goldberg, David E., and John H. Holland. "Genetic algorithms and machine learning." Machine learning 3.2 (1988): 95-99.

\bibitem[Back(2000)]{back}
Bäck, Thomas, David B. Fogel, and Zbigniew Michalewicz, eds. Evolutionary computation
1: basic algorithms and operators. Vol. 1. CRC Press, 2000.

\bibitem[Grefenstette(1992)]{grefenstette}
Grefenstette, John J. "Genetic algorithms for changing environments." PPSN. Vol. 2. 1992.

\bibitem[Cobb(1993)]{cobb}
Cobb, Helen G., and John J. Grefenstette. Genetic algorithms for tracking changing
environments. NAVAL RESEARCH LAB WASHINGTON DC, 1993.

\bibitem[Das(2017)]{das}
Das, P. P., R. Bisoi, and P. K. Dash. "Data decomposition based fast reduced kernel extreme learning machine for currency exchange rate forecasting and trend analysis." Expert Systems with Applications (2017).

\bibitem[Kuroda(2017)]{kuroda}
Kuroda, Kazuma. "Predicting Optimal Trading Actions Using a Genetic Algorithm and Ensemble Method." Intelligent Information Management 9.06 (2017): 229.

\bibitem[Petropoulos(2017)]{petropoulos}
Petropoulos, Anastasios, et al. "A stacked generalization system for automated FOREX portfolio trading." Expert Systems with Applications 90 (2017): 290-302.

\bibitem[de Almeida(2018)]{dealmeida}
de Almeida, Bernardo Jubert, Rui Ferreira Neves, and Nuno Horta. "Combining Support Vector Machine with Genetic Algorithms to Optimize Investments in Forex Markets with High Leverage." Applied Soft Computing (2018).

\bibitem[Sidehabi(2016)]{sidehabi}
Sidehabi, Sitti Wetenriajeng, and Sofyan Tandungan. "Statistical and Machine Learning approach in forex prediction based on empirical data." Computational Intelligence and Cybernetics (CYBERNETICSCOM), 2016 International Conference on. IEEE, 2016.

\bibitem[Deng(2015)]{deng}
Deng, Shangkun, et al. "Hybrid method of multiple kernel learning and genetic algorithm for forecasting short-term foreign exchange rates." Computational Economics 45.1 (2015): 49-89.

\bibitem[Evans(2013)]{evans}
Evans, Cain, Konstantinos Pappas, and Fatos Xhafa. "Utilizing artificial neural networks and genetic algorithms to build an algo-trading model for intra-day foreign exchange speculation." Mathematical and Computer Modelling 58.5-6 (2013): 1249-1266.

\bibitem[Georgios(2013)]{georgios}
Sermpinis, Georgios, et al. "Forecasting foreign exchange rates with adaptive neural networks using radial-basis functions and particle swarm optimization." European Journal of Operational Research 225.3 (2013): 528-540.

\bibitem[Suthaharan(2016)]{suthaharan}
Suthaharan, Shan. Machine learning models and algorithms for big data classification. Boston: Springer, 2016.

\bibitem[Davis(2006]{davis}
Davis, Jesse, and Mark Goadrich. "The relationship between Precision-Recall and ROC curves." Proceedings of the 23rd international conference on Machine learning. ACM, 2006.

\bibitem[Mnih(2015]{mnih}
Mnih, Volodymyr, et al. "Human-level control through deep reinforcement learning." Nature 518.7540 (2015): 529-533.

\end{thebibliography}


\end{document}